%% file: main.tex
\documentclass{article}
\usepackage[T1]{fontenc}
\usepackage{iclr2027_conference,times}
\usepackage{amsmath,amssymb,amsthm,graphicx,booktabs,hyperref,url,placeins}
\hypersetup{hidelinks,pdfauthor={},pdftitle={Hidden, Not Deleted: How Networks Suppress Entangled Features}}
\newtheorem{proposition}{Proposition}
\newtheorem{theorem}{Theorem}
\newcommand{\E}{\mathbb{E}}
\newcommand{\R}{\mathbb{R}}
\newcommand{\relu}{\operatorname{ReLU}}

\title{Hidden, Not Deleted: \\How Networks Suppress Entangled Features}
\author{Akash Samanta\thanks{Corresponding author: \texttt{akashsamanta.web@gmail.com}} \\
Techno India University\\
Salt Lake, Kolkata, India \\
\And
Manish Pratap Singh \\
DRDO Young Scientist Laboratory - CT\\
Chennai, India \\
\And
Debasis Chaudhuri \\
Techno India University\\
Salt Lake, Kolkata, India 
}

\iclrfinalcopy

\begin{document}
\maketitle
\pagestyle{plain}
\thispagestyle{plain}

\maketitle
\raggedbottom
\begin{abstract}
When a neural network stops reporting a feature, has it forgotten the feature or merely hidden it? Shared representations complicate the answer: removing one feature’s direction can also damage another feature. We study this problem in a small autoencoder that compresses and reconstructs inputs, allowing individual feature contributions to be traced directly. With the representation fixed, restoring a single bias partially recovers target detection by exposing a signal hidden behind the output’s activation threshold. Tests on target-absent inputs distinguish this recovery from interventions that simply increase output. The same bias intervention becomes less effective when the representation is allowed to change. We find that training only the other outputs can weaken the target’s contribution, even without a loss on its own output. A conditional theoretical analysis explains how interference with retained predictions creates this pressure. Yet more flexible predictors still detect residual target information. These findings show how recovery depends on feature geometry and on which parameters are trained, and why a silent output is insufficient evidence of forgetting.
\end{abstract}

\section{Introduction}
\label{sec:intro}
Suppose a model reports which attributes are present in an input. We train it to stop reporting attribute B while continuing to report the others, including A. The B output becomes small. Does this mean B has been removed from the model, or can a small change make the model recognize B again? Answering this requires more than making the output increase: a recovered detector must distinguish inputs that contain B from inputs that do not.

Machine unlearning seeks to remove the effects of designated training data or knowledge without training a replacement model from scratch \citep{cao2015towards,bourtoule2021machine}. Recent work shows that unlearning can leave recoverable behavior and internal signals, and studies those signals directly \citep{yang2026erase,xu2026unlearning,jang2026suppression,habbati2026gradient}. Our aim is to explain a specific instance of this problem in a setting where the feature contributions and the parameters that can change are known. We study feature-output suppression, not certified removal of training examples.

\begin{figure}[!htb]
\centering
\includegraphics[width=\linewidth]{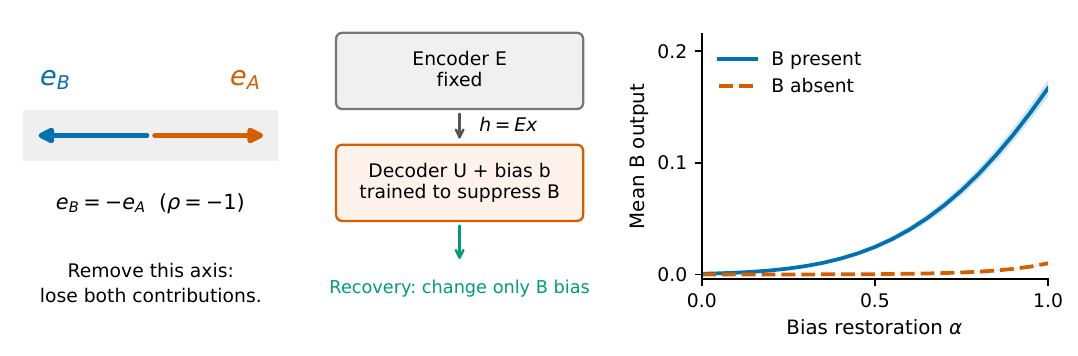}
\par\vspace{2pt}
\makebox[.28\linewidth][c]{\small (a) Shared direction}%
\makebox[.31\linewidth][c]{\small (b) Suppress the output}%
\makebox[.41\linewidth][c]{\small (c) Measured bias recovery}

\caption{Shared geometry, output suppression, and bias recovery. (a) A and B use opposite directions on one axis; removing the axis removes both contributions. (b) With the encoder fixed, suppression trains the decoder and bias. Recovery changes only B's bias. Panels (a--b) are schematic. (c) Measured B output as its bias moves linearly from the suppressed value ($\alpha=0$) to the saved reconstruction reference ($\alpha=1$): mean $\pm$ seed standard error over all 100 antipodal models, evaluated on mixed sparse inputs. B-present and B-absent responses separate; Section~\ref{sec:recovery} quantifies detection and reconstruction errors.}
\label{fig:overview}
\end{figure}

Our model compresses a sparse vector of feature magnitudes into a smaller hidden vector, then reconstructs it. Several features must share this space; the use of overlapping representation directions is called \emph{superposition} \citep{elhage2022toy}. A particularly clear example places A and B in opposite directions on the same axis, an \emph{antipodal pair}. A linear edit that removes B's direction also removes A's. A nonlinear output can instead suppress B while preserving A's output (Figure~\ref{fig:overview}). Whether the suppressed feature remains selectively recoverable is the central question.

We answer three linked questions. First, \emph{what does shared geometry constrain?} We give exact limits on linear contribution removal and compare representation edits with readout suppression on matched inputs. Second, \emph{does a readout patch recover the right feature?} In the antipodal subset of the 1,500-model geometry sweep, restoring the saved bias recovers detection, whereas taking the absolute value of the suppressed score fails on B-absent inputs. The measured effect of B on its own score explains why these interventions differ. Third, \emph{when does this recovery persist?} Paired architecture and encoder-training interventions show that a small output, a recoverable readout, and an attenuated internal contribution are different outcomes.

The fixed-encoder result does not demonstrate learned preservation: fixing the encoder preserves its representation by construction. Its informative part is the selectivity and failure of particular output interventions. Allowing representation-forming parameters to change tests that boundary. A loss ablation then identifies a cause of attenuation: training the other outputs can weaken B even when B's own output has no loss. We support this explanation with an explicitly conditional reconstruction result and test residual information with both linear and nonlinear probes.

The study connects shared geometry to selective recovery through interventions on known feature contributions. It identifies when a suppressed score remains useful, when a bias change cannot recover it, and how the losses of other outputs can weaken the target representation. This provides a mechanistic account of suppression in a model where each of these effects can be measured separately.

\section{Model, interventions, and evaluation}
\label{sec:setup}
\paragraph{A feature has a known meaning in the model.}
Each coordinate of $x\in\R^{16}$ is independently present with probability $0.1$. A present magnitude is uniform on $[0.2,1]$; zero means absence. A and B are coordinates 0 and 1 and can co-occur. The base autoencoder has a four-dimensional hidden state and separate encoder and decoder weights:
\begin{equation}
 h=Ex,\qquad \hat x=\relu(Uh+b),\qquad
 E\in\R^{4\times16},\quad U\in\R^{16\times4}.
 \label{eq:model}
\end{equation}
Write $e_i=E_{:,i}$ and $u_i=U_{i,:}^{\mathsf T}$. The contribution $e_Bx_B$ is directly known. In imposed-geometry experiments, $\|e_A\|=\|e_B\|=0.2$ and their cosine is $\rho$; the other columns use Gaussian initialization. The encoder is fixed throughout pretraining. Separate experiments learn the encoder or add an intermediate layer. ReLU is the base output activation \citep{nair2010rectified}; Section~\ref{sec:extensions} tests GELU.

\paragraph{Suppression objective and paired training.}
After reconstruction pretraining, we minimize
\begin{equation}
 L_\beta=\E\left[\frac1n\sum_{i=1}^n(\hat x_i-x_i)^2+\beta\hat x_B^2\right],
 \qquad n=16,\quad\beta=1.
 \label{eq:loss}
\end{equation}
B's reconstruction term remains present; the added penalty has relative coefficient $n\beta=16$. We call the 15 non-B coordinates the \emph{retained outputs}. Paired experiments use Adam \citep{kingma2015adam}, learning rate $10^{-3}$, no weight decay, and batches of 512: 15,000 pretraining steps, an Adam reset and 3,000 reconstruction warmup steps, then 5,000 intervention steps. Paired branches share starting weights, available optimizer moments, data, and update count. The geometry sweep restores a reference bias saved after pretraining, before warmup. Paired experiments save the reference state immediately before suppression. These reference points are kept distinct throughout the analysis.

\paragraph{Cohorts and controls.}
The geometry sweep contains 1,500 models, with 100 seeds at each of 15 encoder cosines, including $\rho=-1$. Paired base-model comparisons use 30 seeds at each of $\rho\in\{-1,-0.5,0\}$; another 30 learn their encoder without geometry selection. Depth-2 ReLU and depth-1 GELU use 30 seeds each. A 1,200-run emergence sweep measures how often antipodal geometry arises during reconstruction training. Every seed is retained, including poor pretrained models. Appendix~\ref{app:protocol} specifies controls, random streams, and uncertainty.

\paragraph{Recovery requires positive and negative tests.}
The primary test uses 4,096 independent sparse inputs, including A/B co-occurrence and other features. We measure B-presence AUROC, average precision, B-present and B-absent mean squared error (MSE), and retained-output MSE. AUROC measures ranking, not output scale or accurate reconstruction. Each intervention's decision threshold is the 99th percentile of B-absent scores on a separate 8,192-input validation set. We report both the test true-positive rate (TPR) and the actual false-positive rate (FPR); ties can make FPR lower than 1\%. Additional tests isolate A only, B only, both, unrelated features, and the empty input. No patch is chosen using test labels.

A \emph{probe} is a separately trained predictor of B presence from the hidden state. Its training and test inputs are independent. Probe failure does not certify absence of information; smaller column norms do not certify it either. Tables report mean $\pm$ seed standard deviation on shared test inputs; figure bands use seed standard error.

\section{Shared geometry constrains representation edits}
\label{sec:geometry}
Linear concept-erasure methods remove information accessible to specified predictors \citep{ravfogel2020null,ravfogel2022linear,belrose2023leace}. Our starting geometric question is narrower: can a linear map remove B's contribution while preserving A's contribution? Let $u,v$ be unit encoder directions for A and B, with $u^{\mathsf T}v=\rho$. Orthogonal removal along B, $P=I-vv^{\mathsf T}$, leaves $\|Pu\|^2=1-\rho^2$. At antipodality it removes both directions. This is a statement about hidden contributions, not a downstream MSE formula.

\begin{proposition}[Cost of preserving one shared direction]
\label{prop:linear}
If $Pu=u$ and $Pv=0$, then for $|\rho|<1$,
\begin{equation}
 \|P\|_{\rm op}\geq(1-\rho^2)^{-1/2}.
 \label{eq:linearbound}
\end{equation}
The bound is attainable. At $|\rho|=1$, these requirements are incompatible. More generally, if $\|Pu-u\|\leq\delta$, $\|Pv\|\leq\epsilon$, and $\|P\|_{\rm op}\leq K$, then $\delta+|\rho|\epsilon\geq1-K\sqrt{1-\rho^2}$.
\end{proposition}

Thus nearly shared directions permit exact separation by an oblique linear map, but require increasing worst-case amplification. This elementary constraint does not invalidate LEACE: its guarantee concerns linear predictability and minimal distortion, not preservation of every other task. Table~\ref{tab:geometry} compares fitted LEACE, encoder-direction removal, and readout interventions on identical models and inputs. Removing the shared contribution increases population A error. Readout suppression leaves other rows' paired trajectories unchanged because their gradients are independent when $E$ is fixed. Appendix~\ref{app:proofs} proves both facts; Appendix~\ref{app:protocol} also reports decoder-direction projection and output-zeroing controls.
\input{generated/geometry_main.tex}

\paragraph{The objective can reduce magnitude without removing discrimination.}
For B target magnitude $t$ and output $y$,
\begin{equation}
 \frac{(y-t)^2}{n}+\beta y^2
 =\frac{1+n\beta}{n}\left(y-\frac{t}{1+n\beta}\right)^2
 +\frac{\beta t^2}{1+n\beta}.
 \label{eq:square}
\end{equation}
For a fixed encoder, ReLU readouts are closed under positive scaling. Scaling a reconstruction-optimal B readout by $1/17$ therefore gives a minimizer of its penalized problem within that class. Scaling any existing readout by a positive constant preserves its AUROC, as the matched baseline illustrates. Finite Adam training need not reach this optimum: many trained outputs become almost entirely clipped to zero. The next section tests whether their underlying score still supports selective recovery.

\section{Selective recovery after readout suppression}
\label{sec:recovery}
For an isolated B input of magnitude $m$, the suppressed ReLU readout has pre-activation
\begin{equation}
 z_B(m)=q_Bm+b_B,\qquad q_B=u_B^{\mathsf T}e_B.
 \label{eq:gain}
\end{equation}
A positive functional gain $q_B$ means that B still increases its own pre-activation, even when the bias keeps the output near zero. All 100 antipodal models in the geometry sweep have $q_B>0$. This gain directly measures how B affects its score. The angle between two decoder rows measures a different relation and does not determine the sign of $q_B$ (Appendix~\ref{app:endpoints}).

\begin{figure}[!htb]
\centering
\includegraphics[width=\linewidth]{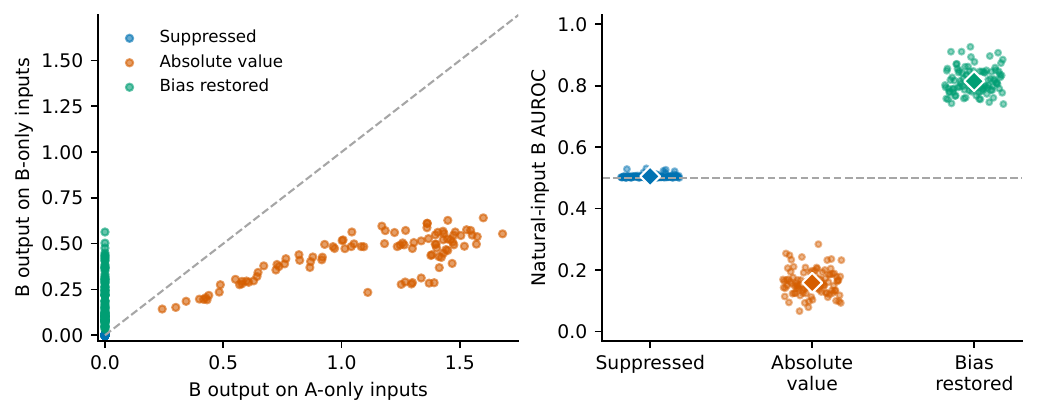}
\caption{Recovery must distinguish B from its negatives. Each point is an antipodal model from the geometry sweep ($N=100$). Left: mean B output on matched isolated inputs; the diagonal marks equal responses to A only and B only. Right: B-presence AUROC on mixed sparse inputs. Bias restoration recovers discrimination; the absolute-value intervention fails these negative controls. Right-panel diamonds mark seed means.}
\label{fig:recovery}
\end{figure}

\paragraph{Bias restoration recovers useful, incomplete discrimination.}
Replacing only $b_B$ with its saved reconstruction reference value increases mean B-presence AUROC from $0.505$ to $0.815$ (Table~\ref{tab:archive}). At independently calibrated thresholds, it detects $35.5\%$ of B-present inputs with $1.09\%$ false positives. B-present MSE remains $0.223$: this is partial selective recovery, not accurate reconstruction of every magnitude. Figure~\ref{fig:recovery} shows the per-model results. The paired 30-seed antipodal cohort, using immediate pre-suppression biases, gives AUROC $0.504\rightarrow0.820$.
\input{generated/archive_table.tex}

\paragraph{The absolute-value patch reports B when only A is present.}
An alternative intervention replaces $\relu(z_B)$ by $|z_B|$, exposing the magnitude of a score clipped by ReLU. At antipodality, A-only input gives $z_B=-q_Bm+b_B$. When both pre-activations are negative and $q_B>0$,
\begin{equation}
 |z_B(\text{B only})|=-b_B-q_Bm
 < -b_B+q_Bm=|z_B(\text{A only})|.
 \label{eq:abs}
\end{equation}
The patch reverses the desired ranking. Its B-only/A-only AUROC is zero in all 100 antipodal models in the sweep, both for the training magnitude range $[0.2,1]$ and for the higher-magnitude diagnostic $[0.5,1]$. Its mixed-input AUROC is $0.159$, and B-absent MSE is $0.912$. The output increase is therefore a false indication of recovery: it comes with larger responses to inputs that lack B.

\paragraph{A bias can expose a score, but cannot fix its ordering.}
For $s=u_B^{\mathsf T}h$ and threshold $\tau\geq0$,
\begin{equation}
 \relu(s+b')>\tau\quad\Longleftrightarrow\quad s>\tau-b'.
 \label{eq:biascriterion}
\end{equation}
Changing the bias can separate observations that ReLU collapsed to zero; it cannot reverse their underlying score order. Perfect separation of finite positive and negative sets is possible exactly when $\max s_{\rm negative}<\min s_{\rm positive}$. Background interference explains why isolated recovery can be stronger than mixed-input recovery. Restoring both B's weight row and bias recovers its pre-suppression function exactly when the representation is fixed; this gives an exact restoration control.

\paragraph{Co-occurrence and interference limit reconstruction.}
Table~\ref{tab:conditions} separates the isolated, background, and both-present tests. Bias restoration reaches $0.946$ isolated AUROC on $[0.2,1]$ magnitudes, but only $0.840$ when B is added to unrelated background features. When both A and B are present, its B MSE is $0.375$. The absolute-value patch even reports a mean B output of $0.766$ on the empty input. These tests expose failures that a positive-input average conceals.

There is also an exact representational obstruction to perfect recovery in the imposed pair. When $e_A=-e_B$, their hidden contribution is $e_B(x_B-x_A)$. Increasing both magnitudes by the same amount leaves this contribution unchanged. Consequently, whenever both inputs remain in the allowed support, two different A/B assignments can yield the same hidden state. For example, an A-only input with magnitude $0.3$ and an A/B input with magnitudes $0.6,0.3$ have identical contributions. No readout of that fixed state can give the correct B value for both inputs. This is a deterministic non-identifiability statement, not a numerical Bayes-error bound. It explains why recovery must be assessed on the actual mixture and why an isolated-input success cannot establish complete feature reconstruction.
\input{generated/recovery_conditions.tex}

\paragraph{The functional explanation is shared across endpoint geometries.}
Decoder-row angles and bias values vary across the 100 antipodal models, but all retain a positive B gain and all fail the isolated absolute-value control. Bias restoration improves mean discrimination across this distribution. The relevant distinction is therefore between exposing a useful signed score and discarding its sign. Appendix~\ref{app:endpoints} relates the measured endpoint parameters to gain and recovery without assuming that parameter clusters identify separate mechanisms.

\section{How far does bias recovery extend?}
\label{sec:extensions}
\subsection{Depth and activation change the answer}
Does the same bias intervention work when the readout is deeper or its activation is smooth? We test two architectures in paired experiments: depth-2 ReLU, $\hat x=\relu(U\relu(MEx+c)+b)$, and depth-1 GELU, $\hat x=\operatorname{GELU}(UEx+b)$ \citep{hendrycks2016gaussian}. Both use the same imposed antipodal encoder and training schedule; $M,c$ remain trainable in the depth-2 model. Table~\ref{tab:architecture} includes all 30 seeds per architecture, without selecting for initial performance.
\input{generated/architecture_main.tex}

In depth-2 ReLU, bias restoration yields only $0.538$ AUROC, versus $0.787$ before suppression. Restoring B's output weights as well reaches $0.688$, still below its pre-suppression performance: fixing the first encoder does not fix the representation reaching the final readout. In GELU, suppression reduces mean output RMS from $0.136$ to $0.007$, yet AUROC remains $0.851$. Restoring the bias increases magnitude without improving ranking. Thus small outputs need not conceal discrimination, and one saved bias need not recover it when internal transformations change. ReLU's exact scaling and clipping arguments are not assumed to apply to GELU. Appendix~\ref{app:architectures} reports false positives, isolated inputs, paired continuation, and restoration controls.

\subsection{Learned geometry and trainable encoders}
\paragraph{Shared directions can emerge, but are not inevitable.}
The emergence sweep trains encoders and measures whether shared directions arise without imposing the pair. Across 1,200 recorded runs varying sparsity, hidden dimension, and a gate that reduces A/B co-occurrence, the fraction with encoder cosine below $-0.9$ rises from $13/240$ without the gate to $116/240$ at its maximum. Among runs with the loss-plateau flag, these fractions are $13/191$ and $113/201$. This supports learned antipodality under the tested synthetic distributions. The gate also changes feature marginals, so it is not a fixed-marginal test of anti-correlation alone (Appendix~\ref{app:emergence}).

\input{generated/emergence.tex}

A separate unselected cohort of 30 learned-encoder models tests recovery without forcing geometry. Mean output AUROC is $0.795$ before suppression, $0.540$ after fixed-encoder suppression, and $0.658$ after bias restoration; all seeds, including poor initial detectors, are included. Restoring row and bias recovers the pre-suppression function. These results broaden the setting while showing that the high recovery in the imposed pair is not universal (Appendix~\ref{app:learned}).

\begin{figure}[!htb]
\centering
\includegraphics[width=\linewidth]{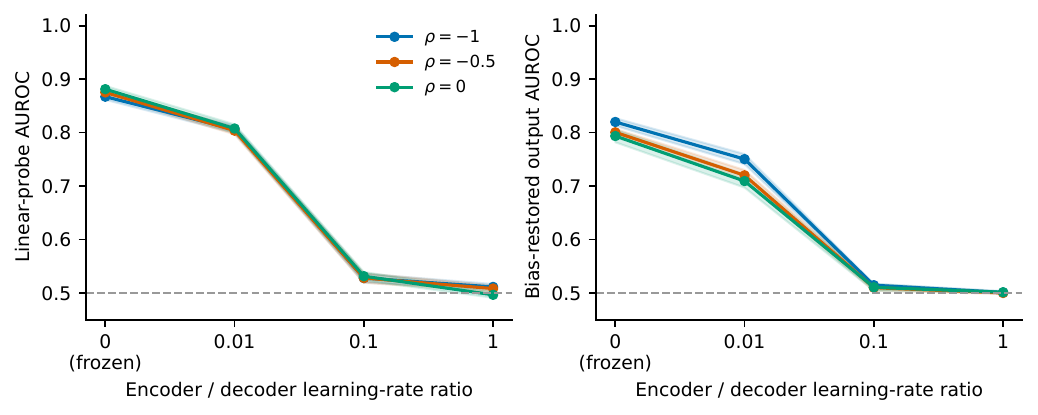}
\caption{Recovery depends on which representation parameters can change. Each curve averages 30 paired seeds per initial encoder cosine. Left: hidden-state linear-probe AUROC. Right: output AUROC after restoring B's saved bias. Bands are seed standard errors. Both shared and orthogonal initial pairs can lose accessible B information under encoder updates.}
\label{fig:adaptation}
\end{figure}

\paragraph{Encoder updates can make bias recovery ineffective.}
We vary the encoder learning rate from zero to $0.01$, $0.1$, or $1$ times the decoder rate in 30 paired models at each initial cosine. At $\rho=-1$ and equal rates, mean $\|e_B\|$ falls from $0.2000$ to $0.0023$; restored-bias AUROC is $0.501$, compared with $0.820$ for the fixed encoder. A logistic-regression probe of the hidden state falls from $0.867$ to $0.511$ AUROC. Figure~\ref{fig:adaptation} also shows non-antipodal geometries. The attenuation trend is not confined to antipodality; shared A/B geometry is not established as a necessary cause.

\section{Why training the other outputs can weaken B}
\label{sec:gradients}
\subsection{A loss ablation identifies the source of attenuation}
Encoder adaptation mixes B's own loss with the losses of the other outputs. To separate them, we start from the 30 antipodal models after fixed-encoder suppression, unfreeze the encoder, reset Adam, and run three paired 5,000-step branches: (i) train only the 15 other outputs with B sampled normally; (ii) use the same loss and minibatches but set input B to zero; (iii) train only B's reconstruction-plus-suppression term. The first two branches leave B's decoder row and bias exactly unchanged. The B-zeroed branch also gives its encoder column exactly zero gradient.

With B still sampled, training the other outputs reduces mean $\|e_B\|$ to $0.0025$ (Table~\ref{tab:gradients}). Setting B to zero leaves it exactly $0.2000$; training B's output alone leaves $0.1756$. Relative to the B-zeroed branch, the B-present branch reduces linear-probe AUROC by $0.194$ (95\% paired seed-bootstrap interval $[0.180,0.209]$). B's own output loss is therefore not necessary for this attenuation. The comparison identifies the effect of B entering the retained reconstruction tasks, not a general claim that retaining unrelated knowledge always deletes a target.
\input{generated/gradient_table.tex}

\begin{figure}[!htb]
\centering
\includegraphics[width=\linewidth]{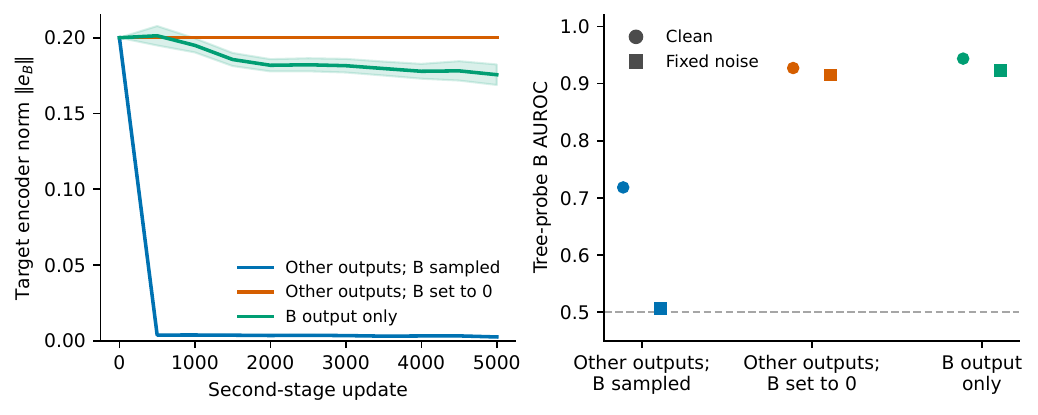}
\caption{Second-stage loss ablation on 30 paired antipodal models. Left: measured B-column norm during training; bands show seed standard error. Right: clean and fixed-noise tree-probe AUROC, with seed standard errors. Training other outputs attenuates B when B remains in their inputs; clean-input nonlinear information survives that attenuation.}
\label{fig:gradients}
\end{figure}

\FloatBarrier
\subsection{Independent interference incurs a reconstruction cost}
B is independent of the retained input coordinates in this experiment. Its contribution can disturb their predictions without helping predict their values. The following result states this explanation under fixed-affine assumptions; Appendix~\ref{app:proofs} supplies the proof and its qualifications.

\noindent\begin{minipage}{\linewidth}
\begin{theorem}[Reconstruction cost of independent interference]
\label{thm:nuisance}
Let $r$ contain the retained features and let $\xi=x_B-\E[x_B]$ be independent of $r$, with variance $0<\sigma_B^2<\infty$. Suppose $\hat r=f_0(r)+C(r)v\xi$, where $f_0,C$ are fixed and independent of $\xi$, and $f_0(r)-r$ and $C(r)$ have finite second moments. With $H=\E[C(r)^{\mathsf T}C(r)]$,
\begin{equation}
 L_R(v)=\frac1n\E\|\hat r-r\|^2
 =L_R(0)+\frac{\sigma_B^2}{n}v^{\mathsf T}Hv.
 \label{eq:nuisance}
\end{equation}
Under gradient flow in $v$, with $f_0,C$ fixed, components in $\ker H$ remain unchanged and components along positive eigenvalues decay.
\end{theorem}
\end{minipage}

Here $v$ is B's encoder contribution direction and $C(r)$ describes its effect on the retained predictions. Components visible to those predictions incur a cost; invisible components do not. For a linear retained decoder, $C=U_R$. For ReLU the result applies within a gating regime independent of $\xi$, with $C(r)=D_R(r)U_R$. Changing gates, correlated features, and simultaneous Adam updates need not follow this fixed-coefficient flow. The theorem explains a conditional mechanism; the paired loss ablation establishes attenuation in the nonlinear experiment.

\subsection{Attenuation is not complete deletion}
\label{sec:nonlinear}
An exploratory gradient-boosted-tree probe, trained and tested on independent inputs with fixed hyperparameters, still detects B after attenuation. Its mean clean-input AUROC is $0.718$ after other-output training and $0.722$ after full encoder adaptation, despite near-chance linear probes. Before suppression it is $0.931$. A failed bias patch or linear probe therefore does not establish deletion.

We additionally add Gaussian noise at 10\% of the pre-suppression hidden-state RMS, using the same absolute scale across paired branches. Tree AUROC becomes $0.506$ and $0.502$ after the two attenuation interventions, versus $0.915$ before suppression. This is a defined stress test, not a model of naturally occurring noise. The joint evidence supports a smaller and less perturbation-robust contribution with surviving clean-input information. Table~\ref{tab:nonlinear} reports the paired probe results, including branches that preserve a stronger signal. Appendix~\ref{app:learned} explains why keeping the B column fixed need not keep every probe score fixed.
\input{generated/nonlinear_table.tex}

\FloatBarrier
\section{Relation to prior work}
\label{sec:related}
\paragraph{Superposition and interpretable features.}
Toy Models of Superposition already establish that sparse features can share directions and that nonlinear readouts filter interference \citep{elhage2022toy}. Our study uses these geometric ingredients to test selective recovery after suppression and the losses that attenuate a target contribution. Sparse-autoencoder studies extract interpretable features from language-model activations \citep{bricken2023monosemanticity,cunningham2023sparse}; our feature identities instead come from the input generator, allowing exact contribution and loss-source interventions. Circuit analyses motivate studying internal computations rather than only aggregate behavior \citep{olah2020zoom,elhage2021framework,olsson2022context}.

\paragraph{Concept erasure is a specified prediction guarantee.}
Geometric debiasing edits embedding directions \citep{bolukbasi2016man}; INLP uses iterative nullspace projection \citep{ravfogel2020null}; R-LACE formulates linear erasure adversarially \citep{ravfogel2022linear}; LEACE gives a closed-form linear eraser with a distortion guarantee \citep{belrose2023leace}. Kernel methods target richer prediction classes \citep{ravfogel2022kernelized,basuroychowdhury2023kram}. These methods specify prediction classes and distortion objectives. Our matched baselines compare the consequences of representation and readout interventions while keeping those different removal criteria explicit. The geometric obstruction concerns preserving a collinear contribution, which these methods do not guarantee.

\paragraph{Recovery and internal changes in unlearning.}
\citet{yang2026erase} already identify suppressive neuron contributions and use attribution-guided regularization. \citet{xu2026unlearning} study reversibility together with representation drift and internal analyses. \citet{jang2026suppression} test restoration through sparse-autoencoder interventions in image classifiers. \citet{habbati2026gradient} use controlled parameter masking to study gradient concentration in class unlearning. Together these studies establish restoration, internal representations, and gradient placement as relevant to unlearning. We add matched selective-recovery and loss-source interventions in a model with exactly known feature contributions. Appendix~\ref{app:prior} compares the interventions and claims directly.

\paragraph{Intervention specificity and external benchmarks.}
Causal mediation, interchange interventions, model editing, and circuit discovery emphasize checking the behavioral role of internal changes \citep{vig2020investigating,geiger2021causal,meng2022rome,wang2022interpretability,conmy2023towards}. Our failed absolute-value patch illustrates why output increase alone is insufficient. TOFU and WMDP evaluate language-model forgetting or capability removal \citep{maini2024tofu,li2024wmdp}; Bias in Bios is a representation-bias benchmark, not a data-deletion benchmark \citep{dearteaga2019bias}. We do not report experiments on them. A small synthetic-transformer experiment measures a bias-induced probability change; its design and the controls needed to interpret that change are given in Appendix~\ref{app:transformer}.

\section{Discussion and conclusion}
\label{sec:conclusion}
The supported phenomenon is selective recovery after output suppression, with a measurable dependence on where learning is allowed. Shared geometry constrains representation edits. A fixed ReLU readout can retain a useful B score behind its activation threshold, and restoring one bias partially exposes it. A smooth activation can retain ranking even at small output magnitude; an adaptable internal representation can make the same bias intervention ineffective. Training the other outputs is sufficient to attenuate B in the tested independent-feature model, while stronger probes still reveal residual information.

These distinctions suggest three checks for interpreting suppression: test target-absent inputs, separate output magnitude from discrimination and reconstruction, and specify which representation parameters can change. Our saved-bias patch is a mechanistic intervention using a previous checkpoint, not a general attack available without one. The model is small, the target is a known input coordinate, and the theory is conditional. Real-model validation and dependent semantic features remain open. The study supplies a controlled explanation and explicit counterexamples; it does not establish that LLM unlearning follows the same dynamics or that an attenuated signal has been completely deleted.

\section*{AI use statement}
Generative AI assistance was used to support code debugging, exploratory brainstorming, preliminary drafting, and language polishing. All experimental designs, theoretical derivations, and analyses were directed and independently verified by the authors, with all numerical results computed directly from executed runs and saved checkpoints. The authors take full responsibility for the final claims, proofs, references, code, and this disclosure. 

\section*{Ethics statement}
The experiments use synthetic inputs and do not involve human subjects or personal data. Recovery interventions require access to model internals and, for bias restoration, a saved earlier parameter. The work examines the reliability of suppression evaluations; it does not demonstrate an attack on a deployed system.

\section*{Reproducibility statement}
The supplement contains checkpoints for the geometry sweep and paired experiments, per-seed measurements, protocols, and programs generating the tables and figures. Appendix~\ref{app:protocol} specifies splits, seeds, optimization, uncertainty, and data availability; Appendix~\ref{app:proofs} gives proofs. The emergence and synthetic-transformer studies have per-run records but no model checkpoints available for independent re-evaluation; their reported results are limited accordingly.

\bibliography{references}
\bibliographystyle{iclr2027_conference}
\clearpage
\appendix
\input{proofs}

\input{protocol}
\input{extensions}

\end{document}

%% file: generated/geometry_main.tex
\begin{table}[!htb]
\centering\small
\setlength{\tabcolsep}{3pt}
\caption{Representation and readout interventions on the same 30 antipodal models and sparse test inputs (mean $\pm$ seed SD). A MSE is population error for coordinate A, not an A-only diagnostic.}
\label{tab:geometry}
\begin{tabular}{lrrr}
\toprule
Intervention & B AUROC & A MSE & Retain MSE \\
\midrule
Before intervention & $0.853\!\pm\!0.035$ & $0.0203\!\pm\!0.0060$ & $0.0228\!\pm\!0.0013$ \\
Remove encoder B direction & $0.499\!\pm\!0.011$ & $0.0457\!\pm\!0.0031$ & $0.0276\!\pm\!0.0013$ \\
Fitted LEACE & $0.502\!\pm\!0.009$ & $0.0440\!\pm\!0.0002$ & $0.0270\!\pm\!0.0012$ \\
Scale B readout by $1/17$ & $0.853\!\pm\!0.035$ & $0.0203\!\pm\!0.0060$ & $0.0228\!\pm\!0.0013$ \\
Train fixed-encoder suppression & $0.504\!\pm\!0.005$ & $0.0203\!\pm\!0.0060$ & $0.0228\!\pm\!0.0013$ \\
\bottomrule
\end{tabular}
\end{table}

%% file: generated/archive_table.tex
\begin{table}[!htb]
\centering
\small
\setlength{\tabcolsep}{3pt}
\caption{Antipodal models in the geometry sweep ($N=100$). Primary sparse-input evaluation, mean $\pm$ seed SD. B$+$ and B$-$ condition on target presence/absence. TPR and FPR use a threshold set on separate validation negatives at nominal 1\% FPR.}
\label{tab:archive}
\begin{tabular}{lrrrrr}
\toprule
Intervention & AUROC & B$+$ MSE & B$-$ MSE & TPR & FPR \\
\midrule
Suppressed & $0.505\!\pm\!0.006$ & $0.397\!\pm\!0.001$ & $0.0000\!\pm\!0.0000$ & $0.010\!\pm\!0.012$ & $0.000\!\pm\!0.000$ \\
Absolute value & $0.159\!\pm\!0.040$ & $0.229\!\pm\!0.084$ & $0.9119\!\pm\!0.5398$ & $0.001\!\pm\!0.002$ & $0.011\!\pm\!0.002$ \\
Bias restored & $0.815\!\pm\!0.041$ & $0.223\!\pm\!0.060$ & $0.0014\!\pm\!0.0007$ & $0.355\!\pm\!0.202$ & $0.011\!\pm\!0.002$ \\
\bottomrule
\end{tabular}
\end{table}

%% file: generated/recovery_conditions.tex
\begin{table}[!htb]
\centering\small
\setlength{\tabcolsep}{3pt}
\caption{Controlled input tests on all 100 antipodal models in the geometry sweep (mean $\pm$ seed SD). Isolated AUROC compares B only with A only; background AUROC compares B plus background with the same background distribution without A or B. Both-present inputs have independent A/B magnitudes.}
\label{tab:conditions}
\begin{tabular}{lrrrr}
\toprule
Intervention & Isolated AUROC & Background AUROC & Both: B MSE & Empty: B output \\
\midrule
Suppressed & $0.503\!\pm\!0.012$ & $0.506\!\pm\!0.007$ & $0.414\!\pm\!0.000$ & $0.000\!\pm\!0.000$ \\
Absolute value & $0.000\!\pm\!0.000$ & $0.135\!\pm\!0.044$ & $0.233\!\pm\!0.107$ & $0.766\!\pm\!0.238$ \\
Bias restored & $0.946\!\pm\!0.060$ & $0.840\!\pm\!0.042$ & $0.375\!\pm\!0.032$ & $0.000\!\pm\!0.002$ \\
\bottomrule
\end{tabular}
\end{table}

%% file: generated/architecture_main.tex
\begin{table}[!htb]
\centering\small
\setlength{\tabcolsep}{3pt}
\caption{Architecture checks: 30 seeds per architecture; mean $\pm$ seed SD. The first encoder is fixed in both. The depth-2 model has a trainable intermediate layer. Output RMS measures magnitude; AUROC measures ranking.}
\label{tab:architecture}
\begin{tabular}{lrrr}
\toprule
Architecture / intervention & B AUROC & B RMS & B$+$ MSE \\
\midrule
Depth-2 ReLU / Before suppression & $0.787\!\pm\!0.120$ & $0.124\!\pm\!0.054$ & $0.190\!\pm\!0.092$ \\
Depth-2 ReLU / Suppressed & $0.509\!\pm\!0.031$ & $0.001\!\pm\!0.002$ & $0.397\!\pm\!0.003$ \\
Depth-2 ReLU / Bias restored & $0.538\!\pm\!0.048$ & $0.014\!\pm\!0.015$ & $0.383\!\pm\!0.024$ \\
Depth-1 GELU / Before suppression & $0.853\!\pm\!0.033$ & $0.136\!\pm\!0.025$ & $0.171\!\pm\!0.044$ \\
Depth-1 GELU / Suppressed & $0.851\!\pm\!0.032$ & $0.007\!\pm\!0.003$ & $0.384\!\pm\!0.004$ \\
Depth-1 GELU / Bias restored & $0.851\!\pm\!0.032$ & $0.039\!\pm\!0.020$ & $0.346\!\pm\!0.023$ \\
\bottomrule
\end{tabular}
\end{table}

%% file: generated/emergence.tex
\begin{table}[!htb]
\centering\small
\setlength{\tabcolsep}{3pt}
\caption{Antipodal encoder geometry in the 1,200-run emergence sweep. The plateau column conditions on the loss-plateau flag; the all-run columns include every run.}
\label{tab:emergence}
\begin{tabular}{lrrrr}
\toprule
Co-occurrence gate & All runs & Rate & Plateau runs & Rate \\
\midrule
0.00 & 13/240 & 5.4\% & 13/191 & 6.8\% \\
0.25 & 61/240 & 25.4\% & 60/203 & 29.6\% \\
0.50 & 76/240 & 31.7\% & 72/205 & 35.1\% \\
0.75 & 89/240 & 37.1\% & 83/205 & 40.5\% \\
1.00 & 116/240 & 48.3\% & 113/201 & 56.2\% \\
\bottomrule
\end{tabular}
\end{table}

%% file: generated/gradient_table.tex
\begin{table}[!htb]
\centering
\small
\setlength{\tabcolsep}{3pt}
\caption{Training after suppression ($N=30$ paired models; mean $\pm$ seed SD). Each branch receives 5,000 updates. The other-output branches exclude the B-output loss and differ only in whether input B is sampled normally or set to zero.}
\label{tab:gradients}
\begin{tabular}{lrrrr}
\toprule
Intervention & $\|e_B\|$ & Linear AUROC & Noisy probe & Retain MSE \\
\midrule
After frozen suppression & $0.2000\!\pm\!0.0000$ & $0.867\!\pm\!0.032$ & $0.866\!\pm\!0.031$ & $0.0228\!\pm\!0.0013$ \\
Other outputs; B sampled & $0.0025\!\pm\!0.0011$ & $0.508\!\pm\!0.033$ & $0.503\!\pm\!0.012$ & $0.0152\!\pm\!0.0007$ \\
Other outputs; B set to 0 & $0.2000\!\pm\!0.0000$ & $0.702\!\pm\!0.032$ & $0.701\!\pm\!0.033$ & $0.0162\!\pm\!0.0010$ \\
B output only & $0.1756\!\pm\!0.0371$ & $0.882\!\pm\!0.066$ & $0.875\!\pm\!0.059$ & $0.0270\!\pm\!0.0050$ \\
\bottomrule
\end{tabular}
\end{table}

%% file: generated/nonlinear_table.tex
\begin{table}[!htb]
\centering
\small
\setlength{\tabcolsep}{3pt}
\caption{A more flexible probe reveals residual information ($N=30$, mean $\pm$ seed SD). Gradient-boosted trees use independent probe training/test sets and fixed hyperparameters. Clean-input recoverability remains above chance after attenuation.}
\label{tab:nonlinear}
\begin{tabular}{lrr}
\toprule
Intervention & Clean tree AUROC & Fixed-noise tree AUROC \\
\midrule
Before suppression / frozen encoder & $0.931\!\pm\!0.015$ & $0.915\!\pm\!0.017$ \\
Full encoder adaptation & $0.722\!\pm\!0.021$ & $0.502\!\pm\!0.013$ \\
Other outputs; B sampled & $0.718\!\pm\!0.019$ & $0.506\!\pm\!0.014$ \\
Other outputs; B set to 0 & $0.927\!\pm\!0.015$ & $0.915\!\pm\!0.016$ \\
B output only & $0.944\!\pm\!0.025$ & $0.923\!\pm\!0.028$ \\
\bottomrule
\end{tabular}
\end{table}

%% file: proofs.tex
\section{Proofs and qualifications}
\label{app:proofs}
\subsection{Linear edits}
For orthogonal removal $P=I-vv^{\mathsf T}$ along unit B direction $v$, the retained unit A direction $u$ satisfies
\begin{equation}
 \|Pu\|^2=1-\rho^2,\qquad u^{\mathsf T}v=\rho.
 \label{eq:projection}
\end{equation}
This describes the norm of an edited direction, not downstream MSE. General linear maps can preserve A exactly while removing B when $|\rho|<1$, but their amplification must grow near collinearity.

These are elementary geometric constraints, not a failure of LEACE's guarantee. LEACE removes linear information about specified labels while minimizing distortion; it does not assume an unrelated task can always be preserved. We fit centered LEACE for binary B presence on 16,384 separate samples. We also distinguish projection along $e_B$ from projection along $u_B$, since these directions need not coincide. The proof follows below; the fitted operator is specified at the end of this appendix.

For $|\rho|<1$, write $u=\rho v+\sqrt{1-\rho^2}s$ with $s\perp v$, $\|s\|=1$. If $Pu=u$ and $Pv=0$, then $Ps=u/\sqrt{1-\rho^2}$, giving the norm bound. It is attained by
\[
 P=\frac{u(u-\rho v)^{\mathsf T}}{1-\rho^2}.
\]
This map sends $u$ to $u$, $v$ to zero, and has rank-one operator norm $(1-\rho^2)^{-1/2}$. At $|\rho|=1$, linearity makes the requirements inconsistent. In the approximate case,
\[
 1-\delta\leq\|Pu\|\leq |\rho|\|Pv\|+\sqrt{1-\rho^2}\|Ps\|
 \leq |\rho|\epsilon+K\sqrt{1-\rho^2}.
\]
Rearrangement proves Proposition~\ref{prop:linear}. The statements constrain contribution removal, $Pe_B=0$, not every statistical definition of concept erasure. If A and B are dependent, preserving all of A may preserve information about B; the independent-input experiments avoid that separate obstruction.

\subsection{Scaled readouts and row independence}
Completing the square gives Eq.~\eqref{eq:square}. Let $\mathcal F_E$ be the class of ReLU-affine readouts on fixed $h=Ex$ and let $c=(1+n\beta)^{-1}>0$. The map $f\mapsto cf$ is a bijection of $\mathcal F_E$ because $\relu(cz)=c\relu(z)$. Substituting $f=cg$ makes minimizing $\E[(f-ct)^2]$ equivalent to minimizing $c^2\E[(g-t)^2]$. Scaling a reconstruction minimizer therefore minimizes the penalized problem. Finite training need not reach either optimum.

For frozen $E$, B's loss gradient with respect to $(u_i,b_i)$ is zero for $i\ne B$. Other reconstruction coordinates do not depend on $(u_B,b_B)$. Elementwise Adam maintains separate coordinate moments. Induction gives equal paired non-B parameters and optimizer states at every step when minibatches, initial states, and learning rates agree and there is no global clipping or cross-coordinate coupling. The implementation satisfies these conditions.

If B's output is zero throughout the support with strictly negative pre-activation, its parameter gradient is zero and its B-loss is $\E[t^2]/n$. The realizable constant output $y=c\E[t]$ lowers this by $(\E[t])^2/[n(1+n\beta)]$ when $\E[t]>0$. A strictly dead region is therefore not globally optimal. This does not imply a finite-step optimizer escapes it; Adam momentum can move parameters temporarily even when gradients become zero.

\subsection{Selective bias recovery}
For $\tau\geq0$, $\relu(s+b')>\tau$ iff $s>\tau-b'$. Bias changes shift a threshold and preserve ordering wherever outputs are positive. On finite positive and negative sets, zero false positives and zero false negatives are jointly attainable iff $\max s_{\rm negative}<\min s_{\rm positive}$. Equation~\eqref{eq:abs} follows from $e_A=-e_B$ and negative pre-activations. It certifies failure of the absolute-value intervention in that regime, not arbitrary patches.

\subsection{Retained-output reconstruction}
Under the assumptions of Theorem~\ref{thm:nuisance}, gradient flow in $v$ with $f_0$ and $C$ fixed gives
\begin{equation}
 v(t)=\exp\left(-\frac{2\sigma_B^2}{n}Ht\right)v(0).
 \label{eq:flow}
\end{equation}
Components in $\ker H$ are unaffected; components with positive eigenvalues decay. If $H\succeq\gamma I$, $\gamma>0$, then $\|v(t)\|\leq e^{-2\sigma_B^2\gamma t/n}\|v(0)\|$.

Write $a(r)=f_0(r)-r$. Then
\[
 \|a(r)+C(r)v\xi\|^2=\|a(r)\|^2+2\xi a(r)^{\mathsf T}C(r)v
 +\xi^2v^{\mathsf T}C(r)^{\mathsf T}C(r)v.
\]
Independence gives $\E[\xi\mid r]=0$ and $\E[\xi^2\mid r]=\sigma_B^2$. Taking expectations proves Eq.~\eqref{eq:nuisance}. Its gradient is $2\sigma_B^2Hv/n$, giving Eq.~\eqref{eq:flow}. Diagonalizing positive-semidefinite $H$ proves the nullspace and decay statements.

For a linear encoder, $Ex=E_Rr+v\xi+v\E[x_B]$. The last term can be absorbed into the fixed retained decoder's intercept, or into pre-activation biases for ReLU. The theorem uses coordinates that keep this absorbed intercept and the affine response fixed; it is not an exact trajectory formula for simultaneous uncentered updates of all parameters. If gates depend on $\xi$, the cross term need not vanish. If B and retained features are correlated, unconditional centering does not remove the conditional mean; reconstruction can reward part of its contribution. These qualifications motivate the gradient ablations.

\subsection{LEACE implementation}
On the separate fitting set let $\mu=\E[h]$, $\Sigma=\operatorname{Cov}(h)$, $c=\operatorname{Cov}(h,\mathbf1\{x_B>0\})$. With $W=\Sigma^{+1/2}$ and $a=Wc/\|Wc\|$, the operator is
\[
 h\mapsto P(h-\mu)+\mu,\qquad P=I-W^+aa^{\mathsf T}W.
\]
Covariance is computed in float64; eigenvalues below $10^{-10}$ times the largest are discarded. We verify $\|Pc\|<10^{-8}$ on the fitting data. This fitting-set check does not guarantee removal of all nonlinear or test-set dependence. Validation and testing use separate data. For decoder-direction projection, the target pre-activation is analytically constant; the implementation enforces that constant to avoid rankings caused solely by floating-point roundoff.

%% file: protocol.tex
\section{Experimental protocol and complete comparisons}
\label{app:protocol}
\paragraph{Cohorts.}
The geometry sweep uses seeds 0--99 at each of 15 cosines from 0 to $-1$. Paired imposed-geometry experiments use seeds 100--129 at each of $-1,-0.5,0$; learned-encoder runs use seeds 200--229. Development pilots use seeds 0--4 in a separate CPU cohort and are excluded from confirmation averages. The gradient ablation is designed after that pilot and evaluated on seeds 100--129 at $\rho=-1$. Every reported seed is included regardless of recovery success. Geometries and sample counts were fixed before the confirmation runs; this is an internal protocol record, not external preregistration.

\paragraph{Random streams and shared data.}
For paired base-model seed $s$, pretraining and warmup use stream $17000+s$; paired suppression uses $27000+s$; gradient ablations use $44000+s$. Evaluation uses 910771, threshold validation uses 88317, and LEACE fitting uses 77371. These streams are distinct from training. Models share test inputs for paired comparisons. Controlled tests have 4,096 samples per condition. B-only and A-only magnitudes are matched; the both-present condition uses independent A/B magnitudes. Background coordinates are independently present with probability $0.1$.

\paragraph{Probes.}
Stream $871919+s$ generates 4,096 probe-training and 8,192 probe-test inputs. Hidden coordinates are standardized with probe-training statistics only. Logistic regression uses $C=1$, maximum 300 iterations, and no hyperparameter selection on the probe test set. The noise probe adds independent Gaussian noise before standardization, at $0.1$ times the pre-suppression hidden RMS on probe-training data. The absolute scale is shared across paired branches. Both probes are restricted attacks, not information-theoretic certificates. The additional tree probe uses 100 boosting iterations, learning rate $0.1$, at most 15 leaves, minimum 20 samples per leaf, $L_2$ regularization 1, and no early stopping. These settings were fixed before inspecting its test scores. The tree-probe analysis is exploratory; its test set is independent of the data used to fit the probe. Independent Gaussian draws use stream $55121+s$. A 10,000-resample paired bootstrap over the 30 seeds, with random seed 7319, supplies the reported gradient-ablation interval; it conditions on the fixed test set.

\paragraph{Optimization.}
The paired experiments use CPU PyTorch, float32 training, and one numerical thread per worker. The geometry sweep uses GPU training. These cohorts are independently trained; numerical reproducibility is assessed within the stated environment. Paired frozen branches restore the same Adam moments. Encoders unfrozen for suppression start with empty optimizer state in imposed-geometry runs; learned-encoder runs restore their existing pre-suppression moments. Adam is reset in the second-stage gradient ablation. There is no weight decay or clipping. The smaller decoder-rate control uses $10^{-4}$ for 5,000 steps and is not interpreted as converged training.

\paragraph{Matched intervention controls.}
Paired branches include reconstruction-only continuation; fixed-encoder suppression; encoder learning rates $0.01$, $0.1$, or $1$ times the decoder rate; bias-only or weight-only updates; a reduced decoder rate; and reconstruction continuation with the encoder unfrozen. Immediate baselines zero B's output, scale its readout by $1/17$, project along the encoder or decoder direction, or apply fitted LEACE. These interventions have different removal criteria and training costs. Table~\ref{tab:paired} reports matched results for the paired antipodal cohort. Restorations use the immediate pre-suppression state; the geometry sweep's reference bias is from before warmup. Every patch and threshold is fixed without test labels.
\input{generated/paired_table.tex}
\input{generated/adaptive_table.tex}

\paragraph{Measurements and uncertainty.}
AUROC uses B output and binary B presence. Average precision retains natural class imbalance. Positive MSE measures magnitude reconstruction conditional on B being present; negative MSE measures spurious output when absent. Retained MSE averages 15 non-B coordinates. A-only MSE is never substituted for population A MSE. Tables give mean $\pm$ seed standard deviation. Per-seed values, validation thresholds, and both test rates accompany the supplement. Figure bands use standard error. Samples within a seed, and the same seeds across geometries, are not counted as independent training replicates.
\input{generated/full_table.tex}

\paragraph{Software and evidence organization.}
Training and tensor evaluation use PyTorch \citep{paszke2019pytorch}; arrays and aggregation use NumPy, pandas, and SciPy \citep{harris2020array,mckinney2010data,virtanen2020scipy}; probes and classification metrics use scikit-learn \citep{pedregosa2011scikit}; figures use Matplotlib \citep{hunter2007matplotlib}. Package versions, code, seed-level CSVs, checkpoint paths, protocols, and file hashes accompany the supplement. Each cohort has its own directory and protocol. The geometry sweep and paired experiments include model checkpoints, allowing direct re-evaluation. The emergence and synthetic-transformer studies provide per-run summary records and generating programs; model states are unavailable for independent evaluation. Their results are limited to the measurements recorded by those programs.

%% file: generated/paired_table.tex
\begin{table}[!htb]
\centering
\small
\setlength{\tabcolsep}{3pt}
\caption{Paired antipodal models ($N=30$); mean $\pm$ seed SD on identical inputs. Immediate edits use the pre-suppression model. Restorations use the suppressed model. Readout and representation edits pursue different removal criteria.}
\label{tab:paired}
\begin{tabular}{lrrr}
\toprule
Intervention & B AUROC & B$+$ MSE & Retain MSE \\
\midrule
Before suppression & $0.853\!\pm\!0.035$ & $0.141\!\pm\!0.044$ & $0.0228\!\pm\!0.0013$ \\
Reconstruction continuation & $0.853\!\pm\!0.036$ & $0.141\!\pm\!0.044$ & $0.0228\!\pm\!0.0013$ \\
Zero target output & $0.500\!\pm\!0.000$ & $0.398\!\pm\!0.000$ & $0.0228\!\pm\!0.0013$ \\
Scale target readout by $1/17$ & $0.853\!\pm\!0.035$ & $0.374\!\pm\!0.007$ & $0.0228\!\pm\!0.0013$ \\
Encoder-direction projection & $0.499\!\pm\!0.011$ & $0.386\!\pm\!0.011$ & $0.0276\!\pm\!0.0013$ \\
Decoder-direction projection & $0.500\!\pm\!0.000$ & $0.398\!\pm\!0.000$ & $0.0296\!\pm\!0.0033$ \\
Fitted LEACE & $0.502\!\pm\!0.009$ & $0.394\!\pm\!0.008$ & $0.0270\!\pm\!0.0012$ \\
Frozen-encoder suppression & $0.504\!\pm\!0.005$ & $0.398\!\pm\!0.001$ & $0.0228\!\pm\!0.0013$ \\
Restore target bias & $0.820\!\pm\!0.039$ & $0.221\!\pm\!0.052$ & $0.0228\!\pm\!0.0013$ \\
Restore target weights & $0.557\!\pm\!0.025$ & $0.376\!\pm\!0.012$ & $0.0228\!\pm\!0.0013$ \\
Restore weights and bias & $0.853\!\pm\!0.035$ & $0.141\!\pm\!0.044$ & $0.0228\!\pm\!0.0013$ \\
\bottomrule
\end{tabular}
\end{table}

%% file: generated/adaptive_table.tex
\begin{table}[!htb]
\centering
\small
\setlength{\tabcolsep}{3pt}
\caption{Encoder adaptation at initial $\rho=-1$ ($N=30$). Mean $\pm$ seed SD. The reconstruction control has no suppression penalty. Probe models use separate training and test inputs; noise has a fixed pre-suppression scale.}
\label{tab:adaptive}
\begin{tabular}{lrrrr}
\toprule
Intervention & $\|e_B\|$ & Linear AUROC & Noisy probe & Retain MSE \\
\midrule
Frozen & $0.2000\!\pm\!0.0000$ & $0.867\!\pm\!0.032$ & $0.866\!\pm\!0.031$ & $0.0228\!\pm\!0.0013$ \\
$0.01\times$ decoder LR & $0.1623\!\pm\!0.0148$ & $0.806\!\pm\!0.037$ & $0.805\!\pm\!0.037$ & $0.0198\!\pm\!0.0009$ \\
$0.1\times$ decoder LR & $0.0126\!\pm\!0.0139$ & $0.528\!\pm\!0.045$ & $0.521\!\pm\!0.034$ & $0.0166\!\pm\!0.0006$ \\
$1\times$ decoder LR & $0.0023\!\pm\!0.0008$ & $0.511\!\pm\!0.030$ & $0.502\!\pm\!0.012$ & $0.0152\!\pm\!0.0006$ \\
Unfrozen reconstruction control & $0.3405\!\pm\!0.3739$ & $0.784\!\pm\!0.123$ & $0.782\!\pm\!0.126$ & $0.0165\!\pm\!0.0009$ \\
\bottomrule
\end{tabular}
\end{table}

%% file: generated/full_table.tex
\input{generated/full_-1.0.tex}
\input{generated/full_-0.5.tex}
\input{generated/full_0.0.tex}

%% file: generated/full_-1.0.tex
\begin{table}[!htb]
\centering
\small
\setlength{\tabcolsep}{2pt}
\caption{Additional primary-distribution metrics and a controlled co-occurrence test at $\rho=-1$ ($N=30$); mean $\pm$ seed SD.}
\label{tab:full-1.0}
\begin{tabular}{lrrrr}
\toprule
Intervention & Average precision & TPR & FPR & Both: B MSE \\
\midrule
Frozen suppression & $0.113\!\pm\!0.009$ & $0.007\!\pm\!0.010$ & $0.000\!\pm\!0.000$ & $0.414\!\pm\!0.000$ \\
Bias restored & $0.558\!\pm\!0.130$ & $0.367\!\pm\!0.174$ & $0.012\!\pm\!0.003$ & $0.378\!\pm\!0.024$ \\
Absolute value & $0.061\!\pm\!0.003$ & $0.001\!\pm\!0.002$ & $0.011\!\pm\!0.002$ & $0.240\!\pm\!0.101$ \\
Train biases only & $0.114\!\pm\!0.005$ & $0.008\!\pm\!0.006$ & $0.000\!\pm\!0.000$ & $0.414\!\pm\!0.000$ \\
Train weights only & $0.290\!\pm\!0.089$ & $0.207\!\pm\!0.093$ & $0.008\!\pm\!0.005$ & $0.413\!\pm\!0.001$ \\
Decoder LR reduced 10-fold & $0.297\!\pm\!0.086$ & $0.227\!\pm\!0.086$ & $0.003\!\pm\!0.003$ & $0.413\!\pm\!0.001$ \\
Full adaptation + bias restored & $0.108\!\pm\!0.002$ & $0.010\!\pm\!0.004$ & $0.011\!\pm\!0.002$ & $0.414\!\pm\!0.000$ \\
\bottomrule
\end{tabular}
\end{table}

%% file: generated/full_-0.5.tex
\begin{table}[!htb]
\centering
\small
\setlength{\tabcolsep}{2pt}
\caption{Additional primary-distribution metrics and a controlled co-occurrence test at $\rho=-0.5$ ($N=30$); mean $\pm$ seed SD.}
\label{tab:full-0.5}
\begin{tabular}{lrrrr}
\toprule
Intervention & Average precision & TPR & FPR & Both: B MSE \\
\midrule
Frozen suppression & $0.112\!\pm\!0.007$ & $0.006\!\pm\!0.008$ & $0.000\!\pm\!0.000$ & $0.414\!\pm\!0.001$ \\
Bias restored & $0.521\!\pm\!0.105$ & $0.310\!\pm\!0.134$ & $0.011\!\pm\!0.002$ & $0.309\!\pm\!0.113$ \\
Absolute value & $0.061\!\pm\!0.003$ & $0.001\!\pm\!0.001$ & $0.011\!\pm\!0.002$ & $0.193\!\pm\!0.070$ \\
Train biases only & $0.113\!\pm\!0.004$ & $0.007\!\pm\!0.004$ & $0.000\!\pm\!0.000$ & $0.414\!\pm\!0.000$ \\
Train weights only & $0.242\!\pm\!0.072$ & $0.150\!\pm\!0.064$ & $0.006\!\pm\!0.004$ & $0.401\!\pm\!0.032$ \\
Decoder LR reduced 10-fold & $0.275\!\pm\!0.075$ & $0.205\!\pm\!0.075$ & $0.004\!\pm\!0.002$ & $0.396\!\pm\!0.041$ \\
Full adaptation + bias restored & $0.108\!\pm\!0.002$ & $0.010\!\pm\!0.004$ & $0.011\!\pm\!0.003$ & $0.412\!\pm\!0.007$ \\
\bottomrule
\end{tabular}
\end{table}

%% file: generated/full_0.0.tex
\begin{table}[!htb]
\centering
\small
\setlength{\tabcolsep}{2pt}
\caption{Additional primary-distribution metrics and a controlled co-occurrence test at $\rho=0$ ($N=30$); mean $\pm$ seed SD.}
\label{tab:full0.0}
\begin{tabular}{lrrrr}
\toprule
Intervention & Average precision & TPR & FPR & Both: B MSE \\
\midrule
Frozen suppression & $0.113\!\pm\!0.007$ & $0.007\!\pm\!0.008$ & $0.000\!\pm\!0.000$ & $0.414\!\pm\!0.001$ \\
Bias restored & $0.496\!\pm\!0.143$ & $0.287\!\pm\!0.158$ & $0.011\!\pm\!0.003$ & $0.246\!\pm\!0.125$ \\
Absolute value & $0.076\!\pm\!0.087$ & $0.008\!\pm\!0.036$ & $0.011\!\pm\!0.001$ & $0.173\!\pm\!0.058$ \\
Train biases only & $0.114\!\pm\!0.006$ & $0.008\!\pm\!0.006$ & $0.000\!\pm\!0.000$ & $0.414\!\pm\!0.000$ \\
Train weights only & $0.229\!\pm\!0.075$ & $0.144\!\pm\!0.071$ & $0.006\!\pm\!0.004$ & $0.397\!\pm\!0.032$ \\
Decoder LR reduced 10-fold & $0.262\!\pm\!0.087$ & $0.193\!\pm\!0.089$ & $0.005\!\pm\!0.003$ & $0.368\!\pm\!0.065$ \\
Full adaptation + bias restored & $0.108\!\pm\!0.002$ & $0.010\!\pm\!0.005$ & $0.011\!\pm\!0.003$ & $0.412\!\pm\!0.007$ \\
\bottomrule
\end{tabular}
\end{table}

%% file: extensions.tex
\FloatBarrier
\section{Endpoint geometry and functional gain}
\label{app:endpoints}
The encoder column $e_B$ describes how input B enters the hidden state; the decoder row $u_B$ describes how that state contributes to B's output. Their inner product $q_B=u_B^{\mathsf T}e_B$ therefore measures B's effect on its own pre-activation. In contrast, $\cos(u_A,u_B)$ compares two output rows. Opposite decoder rows can give positive gain for both features when their encoder columns also point in opposite directions. Decoder-row alignment alone cannot identify whether a suppressed feature has lost its signed response.

\begin{figure}[!htb]
\centering
\includegraphics[width=\linewidth]{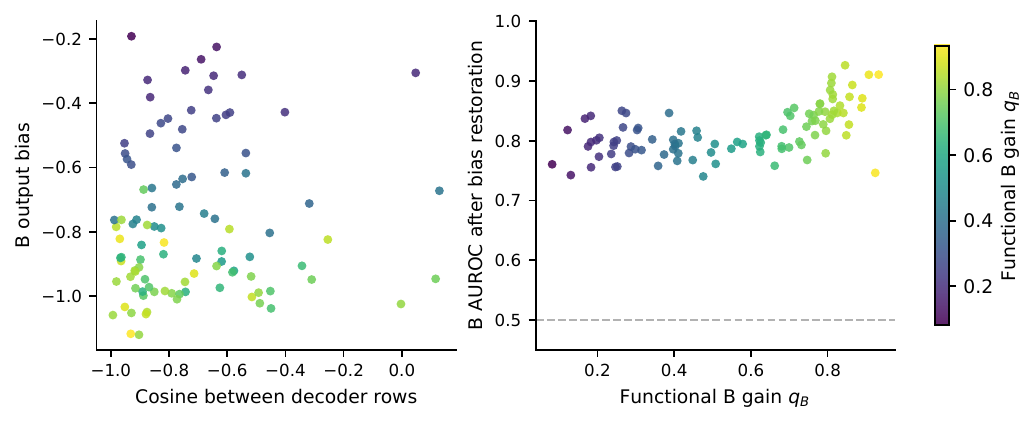}
\caption{Endpoint parameters and functional recovery in the 100 antipodal models. Left: decoder-row cosine and B bias, colored by the positive functional gain $q_B=u_B^{\mathsf T}e_B$. Right: bias-restored B-presence AUROC against that gain, with the same colors. Each point is a measured model. Endpoint coordinates and gain describe the readout; mixed-input recovery additionally depends on background interference and the restored threshold.}
\label{fig:endpoints}
\end{figure}

Figure~\ref{fig:endpoints} shows the endpoint coordinates of all 100 antipodal models and relates their B gains to bias-restored discrimination. Every model has positive B gain. The distribution contains variation in row alignment and bias, yet the same signed-score explanation applies throughout: ReLU can clip a useful score, a bias shift can expose it, and an absolute-value patch can reverse its relation to the target. The population tests in Section~\ref{sec:recovery} measure the resulting discrimination directly.

A strictly negative ReLU pre-activation throughout the input support produces zero gradient for that output. These zero-gradient regions need not be isolated attractors. They are also not globally optimal for Eq.~\eqref{eq:loss} when $\E[x_B]>0$: the constant positive output derived in Appendix~\ref{app:proofs} has lower expected loss. A finite endpoint distribution does not determine the optimizer's global dynamics.

Table~\ref{tab:archive_all} reports all 15 geometries in the sweep. Mixed-input bias recovery also occurs away from antipodality, so recoverable suppression is not confined to a shared A/B axis. The shared axis gives the exact representation-edit obstruction in Section~\ref{sec:geometry}.
\input{generated/archive_all_geometry.tex}

\FloatBarrier
\section{Natural emergence: all recorded runs and denominators}
\label{app:emergence}
The emergence study jointly trains encoder and decoder for 8,000 reconstruction steps with Adam at $10^{-3}$. Its grid is absence probability $S\in\{0.5,0.7,0.8,0.9,0.95,0.99\}$, hidden dimension $d\in\{2,4,8,16\}$, and co-occurrence gate probability $g\in\{0,0.25,0.5,0.75,1\}$, with ten seeds per cell. This gives 1,200 recorded runs. Encoder cosine below $-0.9$ defines an antipodal endpoint.

The generator first samples independent feature presence. If both A and B are present and the gate fires, it removes one of them with equal probability. Writing $p=1-S$, each marginal presence probability becomes $p-gp^2/2$, while joint presence becomes $(1-g)p^2$. Thus changing $g$ changes both marginals and dependence. This is a controlled distributional manipulation, not a pure anti-correlation intervention holding frequencies fixed.

\begin{figure}[!htb]
\centering
\includegraphics[width=\linewidth]{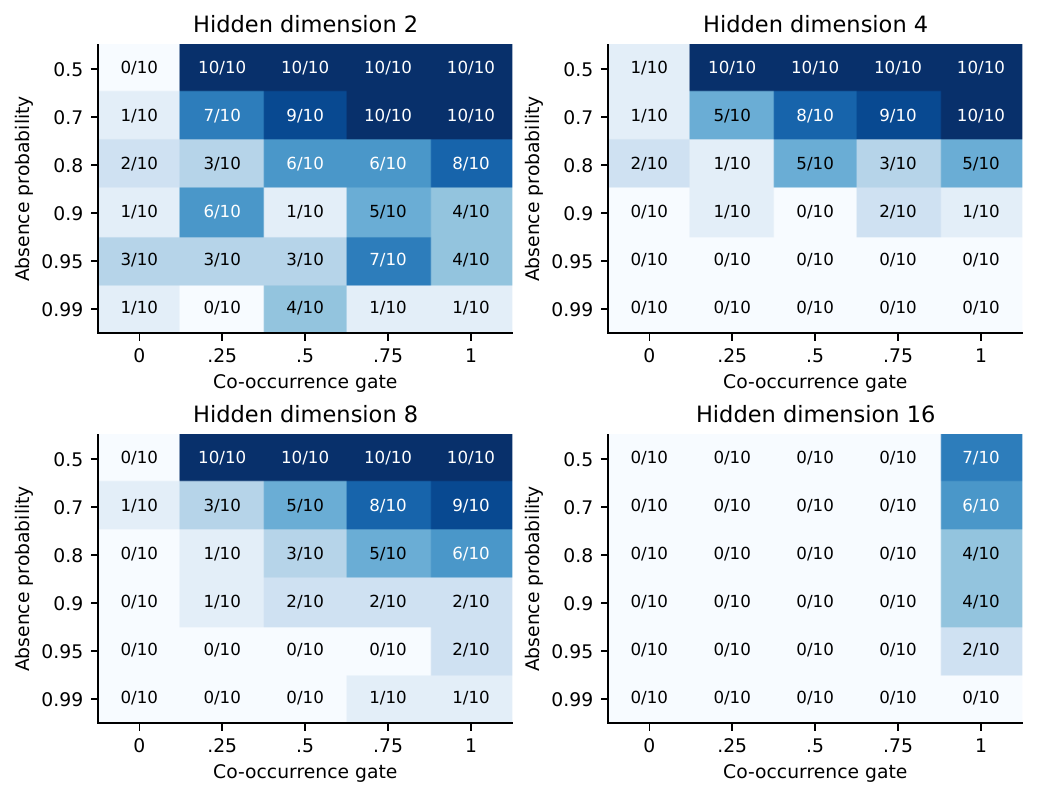}
\caption{Emergence sweep, including every recorded run. Each cell gives the number of encoder cosines below $-0.9$ out of ten seeds. Color represents the same fraction. The geometry can emerge under the tested distributions, but its frequency depends on dimension, sparsity, and the co-occurrence manipulation.}
\label{fig:emergence}
\end{figure}

The loss-plateau diagnostic flags 1,005 runs whose final 10\% of training has less than 1\% relative loss decrease compared with the preceding 10\%. This flag also admits loss increases, so we call it a loss-plateau flag rather than a convergence certificate. Table~\ref{tab:emergence} gives both the all-run result and the result conditional on this flag. Figure~\ref{fig:emergence} gives the full grid, with fixed ten-seed denominators. Per-run endpoint records support these counts; trajectories and model checkpoints are unavailable for independent re-evaluation.

This study supports the availability of shared directions without imposing them in the encoder. It does not establish recovery after suppression: that requires the separate learned-encoder cohort and the explicit recovery measurements below.

\FloatBarrier
\section{Depth and activation: paired recovery tests}
\label{app:architectures}
\paragraph{Design.}
Architecture experiments use seeds 400--429 for each architecture. The first encoder is constructed with $\rho=-1$ and frozen. Decoder and, where present, intermediate weights receive independent Gaussian initialization from the seeded program. The depth-2 model has a four-dimensional intermediate ReLU layer with bias; it is trainable in both reconstruction and suppression branches. GELU replaces the output activation in the depth-1 model and can produce negative outputs. Its reported AUROC uses the signed output; its RMS uses squared outputs.

Training follows Section~\ref{sec:setup}. Pretraining and warmup use stream $61000+s$; the paired 5,000-step branches use $62000+s$. Both branches restore the immediate pre-suppression Adam state. One continues reconstruction only, the other adds B suppression. Each run saves pre-suppression, continuation, and suppressed states. Evaluation and validation use streams 910771 and 88317. The isolated A-only/B-only comparison uses the same 4,096 magnitudes, generated with stream 5713. Protocol and code hashes are saved before execution; this is not external preregistration.

\paragraph{Full controls.}
In depth-2 ReLU, reconstruction continuation has mean B AUROC $0.787\pm0.121$, while restoring both output row and bias after suppression gives $0.688\pm0.110$. The latter intervention cannot undo changes in the intermediate layer. The distinction is therefore between a frozen first encoder and a frozen representation at the final readout. In depth-1 GELU, reconstruction continuation gives $0.853\pm0.033$ AUROC; restoring the output row and bias reproduces the pre-suppression B function exactly. The signed suppression outputs already preserve strong ranking, although magnitude reconstruction is poor. Tables~\ref{tab:architecture_depth2} and~\ref{tab:architecture_gelu} give the remaining tests. All 300 condition-level measurements were recomputed directly from saved tensors using explicit forward equations and an independent rank-based AUROC calculation.
\input{generated/architecture_depth2.tex}
\input{generated/architecture_gelu.tex}

\FloatBarrier
\section{Learned encoders, adaptation, and residual information}
\label{app:learned}
The learned-encoder cohort uses the same 16-to-4 model, but both encoder and decoder are trained during pretraining. Seeds 200--229 are not selected for antipodality, initial B performance, or recovery. A and B remain fixed input coordinates. Table~\ref{tab:learned} includes all seeds and their starting performance. A frozen encoder preserves its hidden state but does not guarantee that restoring one bias reverses changes in the decoder weight row. Restoring row and bias together gives the exact pre-suppression B function.
\input{generated/learned_table.tex}

In this cohort, the tree probe has mean clean AUROC $0.904$ before suppression and $0.777$ after full encoder adaptation. Under the fixed-noise test it changes from $0.753$ to $0.585$. These heterogeneous outcomes reinforce the distinction between attenuation and deletion. The geometry is learned, but the data are still synthetic; this is not a realistic semantic-concept experiment.

Figure~\ref{fig:gradients} reports the second-stage trajectories for the imposed-antipodal cohort and the exploratory nonlinear-probe check. The B-zeroed branch changes other encoder columns even though $e_B$ stays fixed. Consequently, the background distribution in hidden space changes, and linear-probe AUROC need not stay at its starting value. This explains why preserving $\|e_B\|$ is not sufficient to preserve every probe score. Conversely, a very small nonzero $e_B$ can remain recoverable with a sufficiently sensitive nonlinear decoder. No norm threshold is used to declare deletion.

\FloatBarrier
\section{Bias intervention in a synthetic transformer}
\label{app:transformer}
We test an output-bias intervention in a small causal transformer \citep{vaswani2017attention}. The model uses vocabulary size 32, hidden width 32, two blocks, two attention heads, feedforward width 128, sequence length eight, learned token and positional embeddings, and no dropout. Its blocks use ReLU activations. Probes use the first block's hidden state at the last sequence position.

Sequences have A-type, B-type, or neither-type with probabilities $0.4,0.4,0.2$. A-type and B-type sequences do not mix those two tokens. The model uses 6,000 pretraining steps and 2,000 suppression steps with AdamW at $3\times10^{-4}$ \citep{loshchilov2019decoupled}. Suppression freezes embeddings and transformer blocks and updates the unembedding weight and bias. The loss retains next-token cross-entropy at non-B target positions and adds a squared B-probability penalty with coefficient five.

Across 15 recorded runs, the mean A-type and B-type probe accuracies are $0.976$, and mean B probability $0.003061$ before the bias patch and $0.003465$ afterward. These measurements are available as per-run summaries; model checkpoints are unavailable for independent re-evaluation. Thirteen of the fifteen runs meet the pretraining loss-plateau diagnostic. The program also requires probe accuracy above 90\% before proceeding, so this cohort is conditioned on a performance check rather than an unrestricted sample of initializations.

For any fixed logit vector, increasing B's logit bias by $\Delta>0$ changes its softmax probability from $p_B$ to
\[
 p'_B=\frac{e^\Delta p_B}{1+(e^\Delta-1)p_B}>p_B.
\]
A positive probability change is therefore expected algebraically. Without matched target-absent and unrelated-token controls, it does not establish feature-selective restoration. No generation-level recovery, end-to-end unlearning, or transfer to real LLMs is claimed from this experiment.

\clearpage
\section{Comparison with the closest mechanistic predecessors}
\label{app:prior}
Table~\ref{tab:prior} compares the interventions and measurements used in the closest mechanistic studies. It also separates a mechanistic intervention from a practical attack: our bias restoration assumes access to a saved earlier parameter, whereas other studies use different access and training conditions.
\begin{table}[!htb]
\centering\small
\caption{Closest mechanistic predecessors and the narrower contribution of this study. These are scope comparisons, not comparative performance results.}
\label{tab:prior}
\begin{tabular}{p{.19\linewidth}p{.35\linewidth}p{.36\linewidth}}
\toprule
{\raggedright Study\par} & {\raggedright Existing evidence or intervention\par} & {\raggedright Distinction in this study\par} \\
\midrule
{\raggedright \citet{yang2026erase}\par} & {\raggedright Suppressive neuron contributions in LLM unlearning and attribution-guided regularization\par} & {\raggedright Known input directions allow exact contribution accounting and matched positive/negative recovery tests.\par} \\
\addlinespace
{\raggedright \citet{xu2026unlearning}\par} & {\raggedright Reversibility, internal representation drift, and analyses of relearning\par} & {\raggedright We isolate encoder trainability and retained-output gradients to test when a suppressed contribution remains recoverable.\par} \\
\addlinespace
{\raggedright \citet{jang2026suppression}\par} & {\raggedright Representation restoration with sparse-autoencoder interventions in class unlearning\par} & {\raggedright Our saved-bias intervention is simpler and has explicit score-order conditions; it is tested in a known synthetic feature space.\par} \\
\addlinespace
{\raggedright \citet{habbati2026gradient}\par} & {\raggedright Controlled parameter masks and gradient concentration in class unlearning\par} & {\raggedright Our ablation changes the source of the loss and the presence of B in the inputs, exposing a retained-reconstruction route to attenuation.\par} \\
\bottomrule
\end{tabular}
\end{table}

The model does not establish that the same mechanism explains these studies' outcomes. An external test would need a well-performing pretrained model, an explicit forget/retain objective, matched parameter-access conditions, target-present and target-absent evaluations, collateral-damage measurements, and restoration or relearning controls. Such a study is outside the completed experiments and should not be inferred from the synthetic-transformer summary.

%% file: generated/archive_all_geometry.tex
\begin{table}[!htb]
\centering\small
\setlength{\tabcolsep}{3pt}
\caption{The full geometry sweep, with 100 checkpoints per row. Mean $\pm$ seed SD on identical sparse test inputs.}
\label{tab:archive_all}
\begin{tabular}{lrrrr}
\toprule
Initial cosine & Suppressed AUROC & Bias AUROC & Abs. AUROC & Bias B$-$ MSE \\
\midrule
-1.000 & $0.505\!\pm\!0.006$ & $0.815\!\pm\!0.041$ & $0.159\!\pm\!0.040$ & $0.0014\!\pm\!0.0007$ \\
-0.750 & $0.504\!\pm\!0.006$ & $0.815\!\pm\!0.038$ & $0.150\!\pm\!0.041$ & $0.0014\!\pm\!0.0007$ \\
-0.727 & $0.504\!\pm\!0.006$ & $0.816\!\pm\!0.038$ & $0.150\!\pm\!0.041$ & $0.0014\!\pm\!0.0007$ \\
-0.705 & $0.504\!\pm\!0.005$ & $0.816\!\pm\!0.038$ & $0.149\!\pm\!0.041$ & $0.0014\!\pm\!0.0007$ \\
-0.682 & $0.504\!\pm\!0.005$ & $0.817\!\pm\!0.038$ & $0.149\!\pm\!0.040$ & $0.0014\!\pm\!0.0007$ \\
-0.659 & $0.504\!\pm\!0.005$ & $0.814\!\pm\!0.049$ & $0.154\!\pm\!0.071$ & $0.0014\!\pm\!0.0007$ \\
-0.636 & $0.504\!\pm\!0.005$ & $0.814\!\pm\!0.049$ & $0.154\!\pm\!0.073$ & $0.0014\!\pm\!0.0007$ \\
-0.614 & $0.504\!\pm\!0.005$ & $0.814\!\pm\!0.049$ & $0.154\!\pm\!0.073$ & $0.0014\!\pm\!0.0007$ \\
-0.591 & $0.504\!\pm\!0.005$ & $0.814\!\pm\!0.049$ & $0.154\!\pm\!0.075$ & $0.0014\!\pm\!0.0007$ \\
-0.568 & $0.504\!\pm\!0.005$ & $0.814\!\pm\!0.049$ & $0.153\!\pm\!0.075$ & $0.0014\!\pm\!0.0007$ \\
-0.545 & $0.504\!\pm\!0.005$ & $0.811\!\pm\!0.058$ & $0.159\!\pm\!0.099$ & $0.0014\!\pm\!0.0007$ \\
-0.523 & $0.504\!\pm\!0.005$ & $0.811\!\pm\!0.058$ & $0.159\!\pm\!0.099$ & $0.0014\!\pm\!0.0007$ \\
-0.500 & $0.504\!\pm\!0.005$ & $0.811\!\pm\!0.057$ & $0.158\!\pm\!0.099$ & $0.0013\!\pm\!0.0007$ \\
-0.250 & $0.504\!\pm\!0.005$ & $0.810\!\pm\!0.057$ & $0.158\!\pm\!0.104$ & $0.0013\!\pm\!0.0007$ \\
0.000 & $0.504\!\pm\!0.005$ & $0.811\!\pm\!0.048$ & $0.151\!\pm\!0.084$ & $0.0014\!\pm\!0.0007$ \\
\bottomrule
\end{tabular}
\end{table}

%% file: generated/architecture_depth2.tex
\begin{table}[!htb]
\centering\small
\setlength{\tabcolsep}{3pt}
\caption{Depth-2 ReLU: remaining recovery controls ($N=30$, mean $\pm$ seed SD). Thresholds use separate validation negatives. Isolated AUROC compares B-only with A-only inputs.}
\label{tab:architecture_depth2}
\begin{tabular}{lrrrrr}
\toprule
Intervention & B$-$ MSE & Retain MSE & TPR & FPR & Isol. AUROC \\
\midrule
Before suppression & $0.0049\!\pm\!0.0033$ & $0.0243\!\pm\!0.0022$ & $0.300\!\pm\!0.212$ & $0.010\!\pm\!0.004$ & $0.919\!\pm\!0.170$ \\
Reconstruction control & $0.0049\!\pm\!0.0032$ & $0.0242\!\pm\!0.0022$ & $0.301\!\pm\!0.213$ & $0.010\!\pm\!0.004$ & $0.919\!\pm\!0.169$ \\
Suppressed & $0.0000\!\pm\!0.0000$ & $0.0242\!\pm\!0.0022$ & $0.014\!\pm\!0.046$ & $0.000\!\pm\!0.002$ & $0.517\!\pm\!0.071$ \\
Bias restored & $0.0002\!\pm\!0.0007$ & $0.0242\!\pm\!0.0022$ & $0.064\!\pm\!0.062$ & $0.003\!\pm\!0.004$ & $0.549\!\pm\!0.107$ \\
Row + bias restored & $0.0024\!\pm\!0.0051$ & $0.0242\!\pm\!0.0022$ & $0.204\!\pm\!0.153$ & $0.008\!\pm\!0.005$ & $0.767\!\pm\!0.168$ \\
\bottomrule
\end{tabular}
\end{table}

%% file: generated/architecture_gelu.tex
\begin{table}[!htb]
\centering\small
\setlength{\tabcolsep}{3pt}
\caption{Depth-1 GELU: remaining recovery controls ($N=30$, mean $\pm$ seed SD). Thresholds use separate validation negatives. Isolated AUROC compares B-only with A-only inputs.}
\label{tab:architecture_gelu}
\begin{tabular}{lrrrrr}
\toprule
Intervention & B$-$ MSE & Retain MSE & TPR & FPR & Isol. AUROC \\
\midrule
Before suppression & $0.0100\!\pm\!0.0011$ & $0.0260\!\pm\!0.0007$ & $0.303\!\pm\!0.197$ & $0.011\!\pm\!0.002$ & $1.000\!\pm\!0.000$ \\
Reconstruction control & $0.0100\!\pm\!0.0011$ & $0.0260\!\pm\!0.0007$ & $0.303\!\pm\!0.197$ & $0.011\!\pm\!0.002$ & $1.000\!\pm\!0.000$ \\
Suppressed & $0.0000\!\pm\!0.0001$ & $0.0260\!\pm\!0.0007$ & $0.246\!\pm\!0.150$ & $0.011\!\pm\!0.003$ & $1.000\!\pm\!0.000$ \\
Bias restored & $0.0018\!\pm\!0.0019$ & $0.0260\!\pm\!0.0007$ & $0.246\!\pm\!0.150$ & $0.011\!\pm\!0.003$ & $1.000\!\pm\!0.000$ \\
Row + bias restored & $0.0100\!\pm\!0.0011$ & $0.0260\!\pm\!0.0007$ & $0.303\!\pm\!0.197$ & $0.011\!\pm\!0.002$ & $1.000\!\pm\!0.000$ \\
\bottomrule
\end{tabular}
\end{table}

%% file: generated/learned_table.tex
\begin{table}[!htb]
\centering
\small
\setlength{\tabcolsep}{3pt}
\caption{Unselected learned-encoder cohort ($N=30$), mean $\pm$ seed SD. Every seed is included, even when the pretrained model reconstructs B poorly.}
\label{tab:learned}
\begin{tabular}{lrrr}
\toprule
Intervention & B AUROC & B$+$ MSE & Retain MSE \\
\midrule
Before suppression & $0.795\!\pm\!0.168$ & $0.149\!\pm\!0.109$ & $0.0150\!\pm\!0.0010$ \\
Frozen suppression & $0.540\!\pm\!0.090$ & $0.392\!\pm\!0.008$ & $0.0150\!\pm\!0.0010$ \\
Frozen + bias restoration & $0.658\!\pm\!0.128$ & $0.295\!\pm\!0.074$ & $0.0150\!\pm\!0.0010$ \\
Frozen + weight restoration & $0.705\!\pm\!0.181$ & $0.253\!\pm\!0.142$ & $0.0150\!\pm\!0.0010$ \\
Frozen + both restored & $0.795\!\pm\!0.168$ & $0.149\!\pm\!0.109$ & $0.0150\!\pm\!0.0010$ \\
Full adaptation & $0.546\!\pm\!0.112$ & $0.394\!\pm\!0.010$ & $0.0146\!\pm\!0.0011$ \\
Full adaptation + bias restored & $0.519\!\pm\!0.039$ & $0.386\!\pm\!0.014$ & $0.0146\!\pm\!0.0011$ \\
Unfrozen reconstruction control & $0.799\!\pm\!0.163$ & $0.149\!\pm\!0.110$ & $0.0149\!\pm\!0.0010$ \\
\bottomrule
\end{tabular}
\end{table}